\documentclass[runningheads]{llncs}
\usepackage{graphicx}

\usepackage{tikz}
\usepackage{comment}
\usepackage{amsmath,amssymb} 
\usepackage{color}

\usepackage[accsupp]{axessibility}  

\begin{document}
\pagestyle{headings}
\mainmatter
\def\ECCVSubNumber{6}  

\title{MEEV: Body Mesh Estimation On Egocentric Video} 

\titlerunning{MEEV: Body Mesh Estimation On Egocentric Video}
%
\author{Nicolas Monet\inst{1} \and Dongyoon Wee\inst{1}}
\authorrunning{N. Monet et al.}
%
\institute{NAVER CLOVA Video}

\maketitle

\begin{abstract}
This technical report introduces our solution, MEEV, proposed to the EgoBody Challenge at ECCV 2022. Captured from head-mounted devices, the dataset consists of human body shape and motion of interacting people. 
The EgoBody dataset has challenges such as occluded body or blurry image. In order to overcome the challenges, MEEV is designed to exploit multiscale features for rich spatial information. Besides, to overcome the limited size of dataset, the model is pre-trained with the dataset aggregated 2D and 3D pose estimation datasets.
Achieving 82.30 for MPJPE and 92.93 for MPVPE, MEEV has won the EgoBody Challenge at ECCV 2022, which shows the effectiveness of the proposed method. 
The code is available at \url{https://github.com/clovaai/meev}

\end{abstract}

\section{Introduction}
The EgoBody Challenge\cite{Zhang:ECCV:2022} deals with the 3D human pose and shape estimation task from  monocular RGB image. Specifically, they propose a dataset with images captured from egocentric views, which makes challenging problems such as high occlusion and severe motion blur compared to the other datasets used in pose or shape estimation task, e.g., 3DPW \cite{vonMarcard2018}.

In order to tackle the challenging task, we proposed MEEV, which exploit multi-scale features from image encoder and pre-training with large datasets, e.g., AGORA \cite{Patel:CVPR:2021}

In the first section, we describe our approach and then details the experiments we made with our results.

\section{Our Approach}
Extending from the work of Moon et al. \cite{Moon_2022_CVPRW_Hand4Whole}, MEEV additionally use the feature from higher resolution to supplement spatial information. 

\begin{figure}[h]
    \centering
    \includegraphics[width=122mm,scale=0.8]{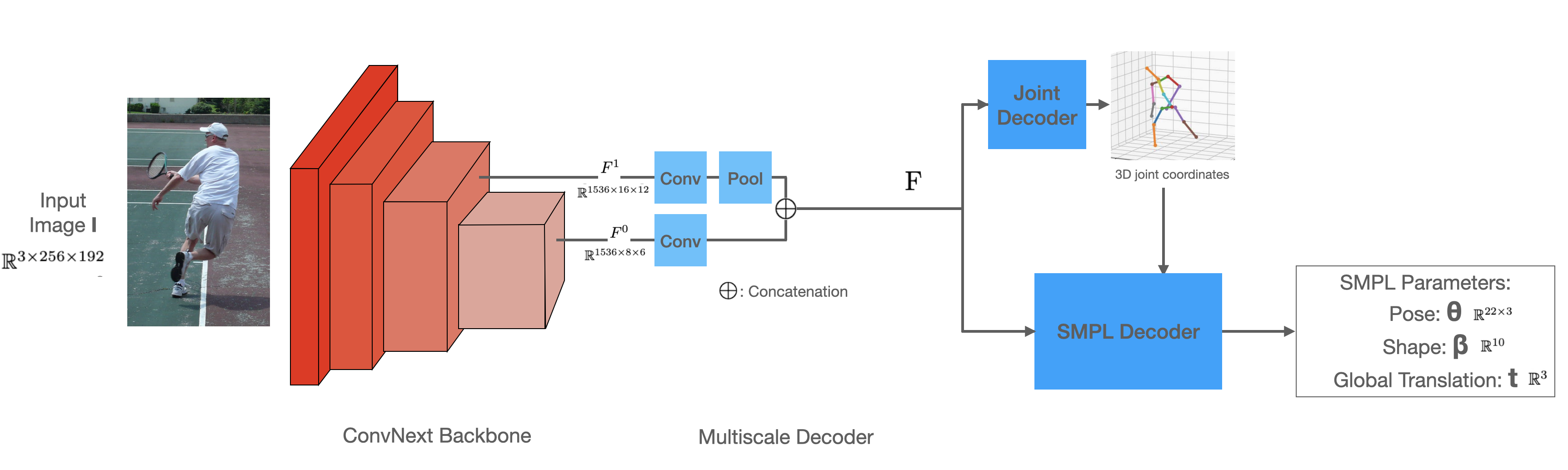}
    \caption{The overview of the whole network architecture.}
    \label{fig:main_model}
\end{figure}

The input image resolution $I \in \mathbb{R}^{3 \times 256 \times 192} $ is cropped from the image using the bounding box of the person.
We use a ConvNext \cite{liu2022convnet} large backbone for the input image.
As shown in Fig. \ref{fig:main_model}, the image features ${F}^0 \in \mathbb{R}^{1536 \times 8 \times 6} $ and ${F}^1  \in \mathbb{R}^{1536 \times 16 \times 12}$
 of different resolution are used to help the 3D decoder to keep some spatial information. Then, 
$F^1$ images features are pipe into a convolution and pool layer to be concatenate with $F^0$ image features to \textbf{F}.
From this image feature \textbf{F}, we generate a 3D heatmap \textbf{H} of human joints. We get the 3D joint coordinates \textbf{C} by using a differentiable soft-argmax on \textbf{H}. 
These coordinates \textbf{C} are used with image features \textbf{F} in the SMPL decoder to predict the SMPL parameters like in Moon et al \cite{Moon_2022_CVPRW_Hand4Whole}.
The final output of the model are the following SMPL \cite{SMPL:2015} body parameters: pose $\theta \in \mathbb{R}^{22 \times 3} $
, shape $\beta \in \mathbb{R}^{10}$ and global translation $t \in \mathbb{R}^{3}$ .

\section{Experiments}
Here, we will first describe the experimental details and then explain the results of our method.

\subsection{Experimental Details}

We first train the model on the following datasets: Human 3.6M \cite{h36m_pami}, MPI INF \cite{mono-3dhp2017}, MSCOCO \cite{mscoco}, AGORA \cite{Patel:CVPR:2021} and MPII \cite{andriluka14cvpr}. The model are initialized with ImageNet pretrained weight.
A basic augmentation (color, affine, blur, coarse dropout, ... ) is apply randomly on each images.
The pretraining use 1e-4 for initial learning rate and a decay after 10 and 20 epochs. We stop after 25 epochs.

The model is then fine-tuned on EgoBody \cite{Zhang:ECCV:2022} dataset for 20 epochs. The learning rate is set to 1e-5 with a batch-size of 48. We use Adam as optimizer for pretraining and fine-tuning. The same augmentation method is used for both training stages.

\subsection{Experimental Results}

We evaluate our method on 3DPW \cite{vonMarcard2018} and EgoBody \cite{Zhang:ECCV:2022} datasets as shown in Table 1.
For the metrics, mean per joint position error (MPJPE) is used as a metric to evaluate 3D joint by calculating the average 3D joint distance (mm) between the predicted and GT, after aligning a root joint translation. Identically, mean per-vertex position error (MPVPE) is used as a metric for measuring the mesh vertices positions by calculating the 3D mesh vertex distance (mm). We use the pelvis as the root joint to calculate the 3D errors.

In Table 1, we compare the performance according to the structure of backbone and decoder. 
The first row is the baseline using Moon et al. \cite{Moon_2022_CVPRW_Hand4Whole} architecture that use a resnet50 \cite{resnet} backbone and the PositionNet as decoder. For the second row and third row, we respectively change the backbone to HRNet W32 \cite{hrnet} and ConvNext \cite{liu2022convnet} large while keeping the same PositionNet decoder.
Finally, in the last row, a ConvNext \cite{liu2022convnet} large backbone with our Multiscale decoder show the best performance.

\setlength{\tabcolsep}{4pt}
\begin{table}
\begin{center}
\begin{tabular}{llllll}
\hline\noalign{\smallskip}
Backbone  & Decoder & MPJPE & MPJPE & MPVPE \\
  &  &(3DPW) & EgoBody & EgoBody \\
\noalign{\smallskip}
\hline
\noalign{\smallskip}
Resnet 50 \cite{resnet} &  PositionNet \cite{Moon_2022_CVPRW_Hand4Whole} & 86.61 & 92.66 & 105.33 \\
HRNet W32 \cite{hrnet} & PositionNet \cite{Moon_2022_CVPRW_Hand4Whole} & 83.37 & 87.42 & 98.21 \\
ConvNext \cite{liu2022convnet} &  PositionNet \cite{Moon_2022_CVPRW_Hand4Whole} & 82.09 & 86.08 & 96.4 \\
\textbf{ConvNext \cite{liu2022convnet} } & \textbf{MultiScaleNet} & \textbf{81.74} & \textbf{82.30} & \textbf{92.93} \\
\hline
\end{tabular}
\caption{Details results on 3DPW \cite{vonMarcard2018} and EgoBody \cite{Zhang:ECCV:2022} datasets for MPJPE and MPVPE metrics using different variants of the model. First line is Moon et al. \cite{Moon_2022_CVPRW_Hand4Whole} paper. }
\end{center}
\label{table:results}
\end{table}
\setlength{\tabcolsep}{1pt}

\subsection{Visual Results}

In figure 2, we compare our results to ground truth for given images. Original image, the predicted result of MEEV and ground truth are shown sequentially in column. Second and third rows show the result in the case of high occlusion whereas, last row show the result in a blurry image case. Though the predicted results have discrepancy compared to the corresponding ground truth, MEEV demonstrate the robust performance on severe conditions, e.g., occluded body, blurry image.

\begin{figure}[h!]
\setlength{\tabcolsep}{5pt}
\setlength{\arrayrulewidth}{0mm}
\renewcommand{\arraystretch}{2}
\begin{tabular}{|c|c|c|}
\includegraphics[width=35mm,scale=0.5]{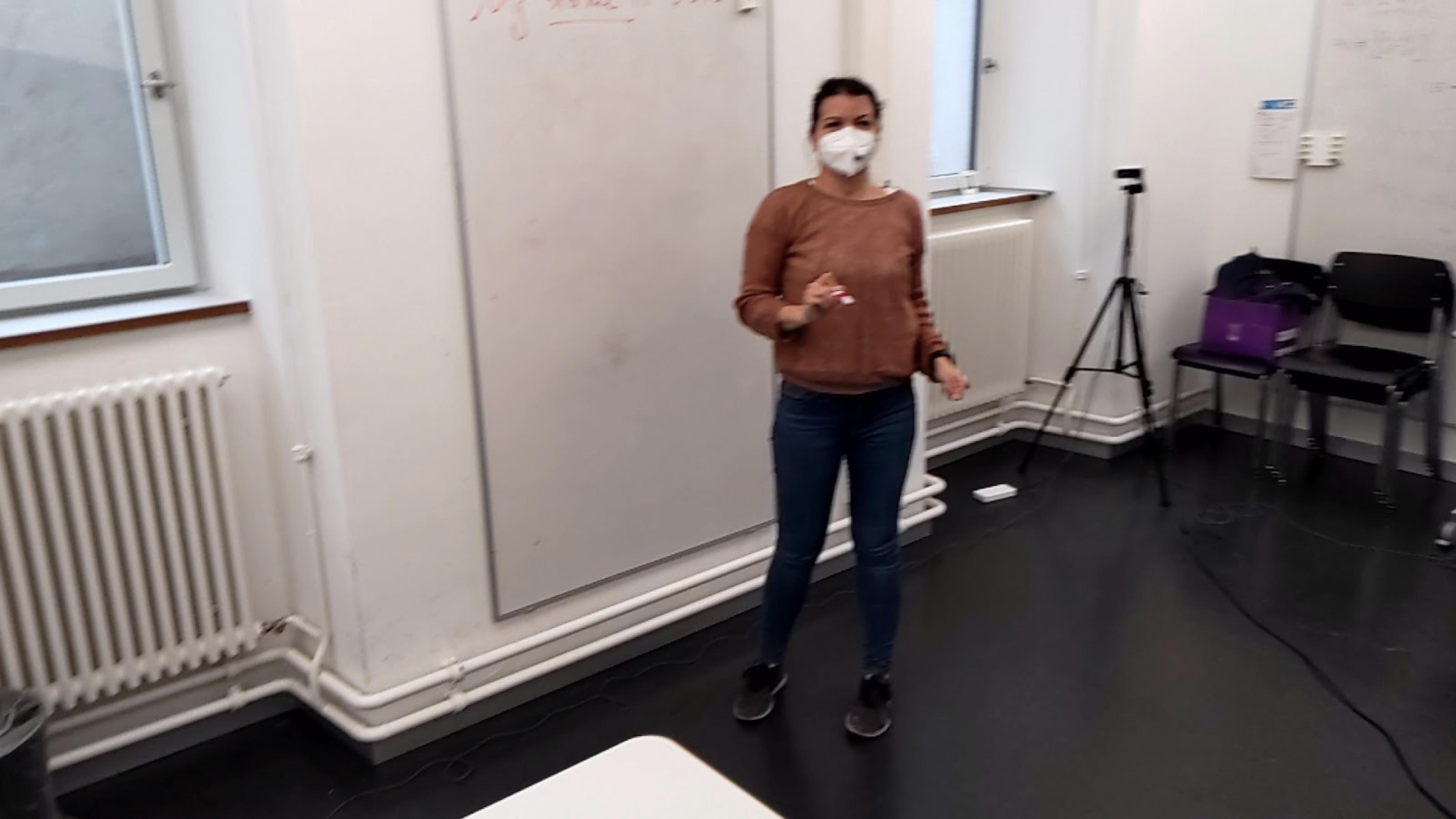} & \includegraphics[width=35mm,scale=0.5]{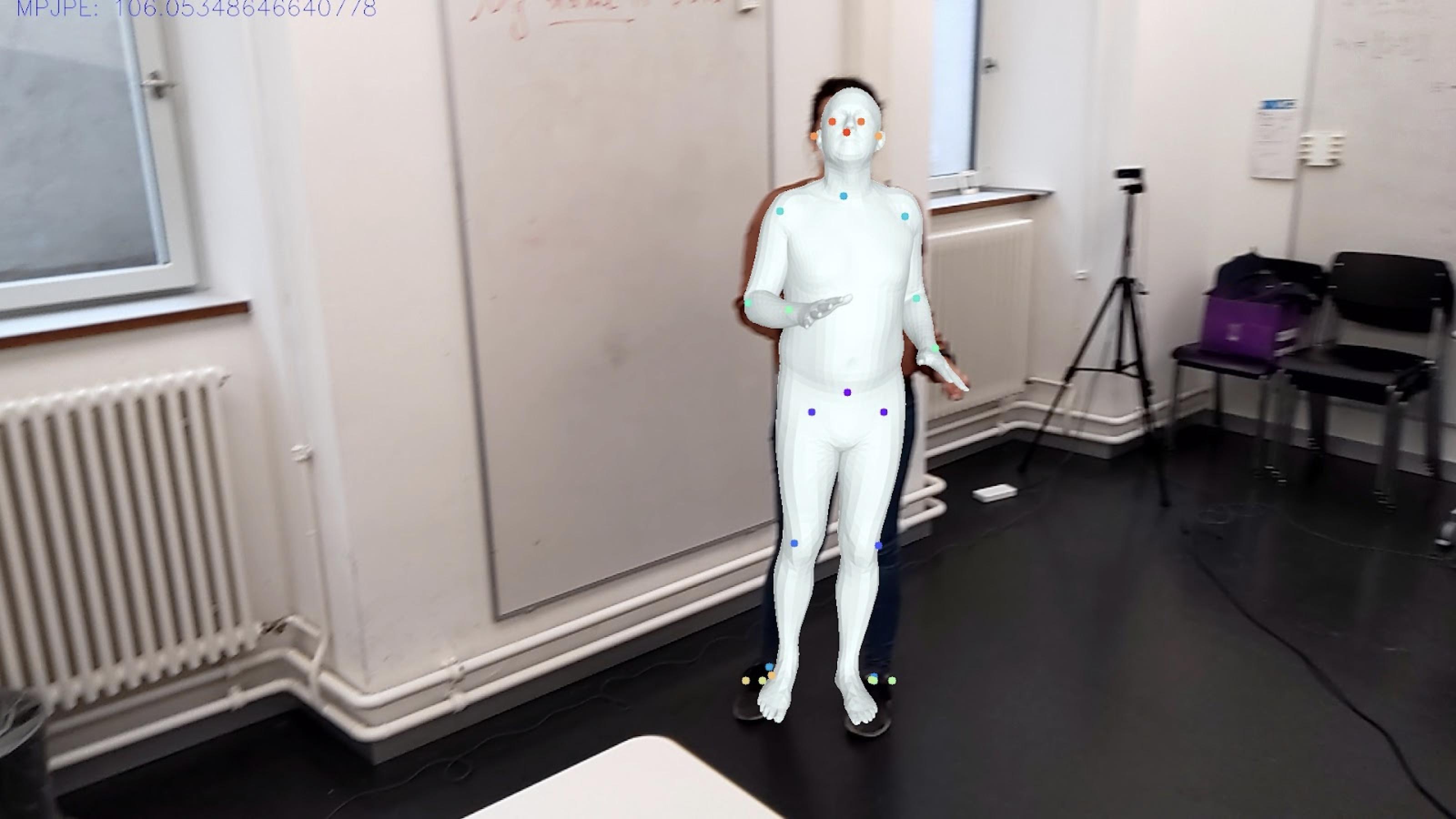} & \includegraphics[width=35mm,scale=0.5]{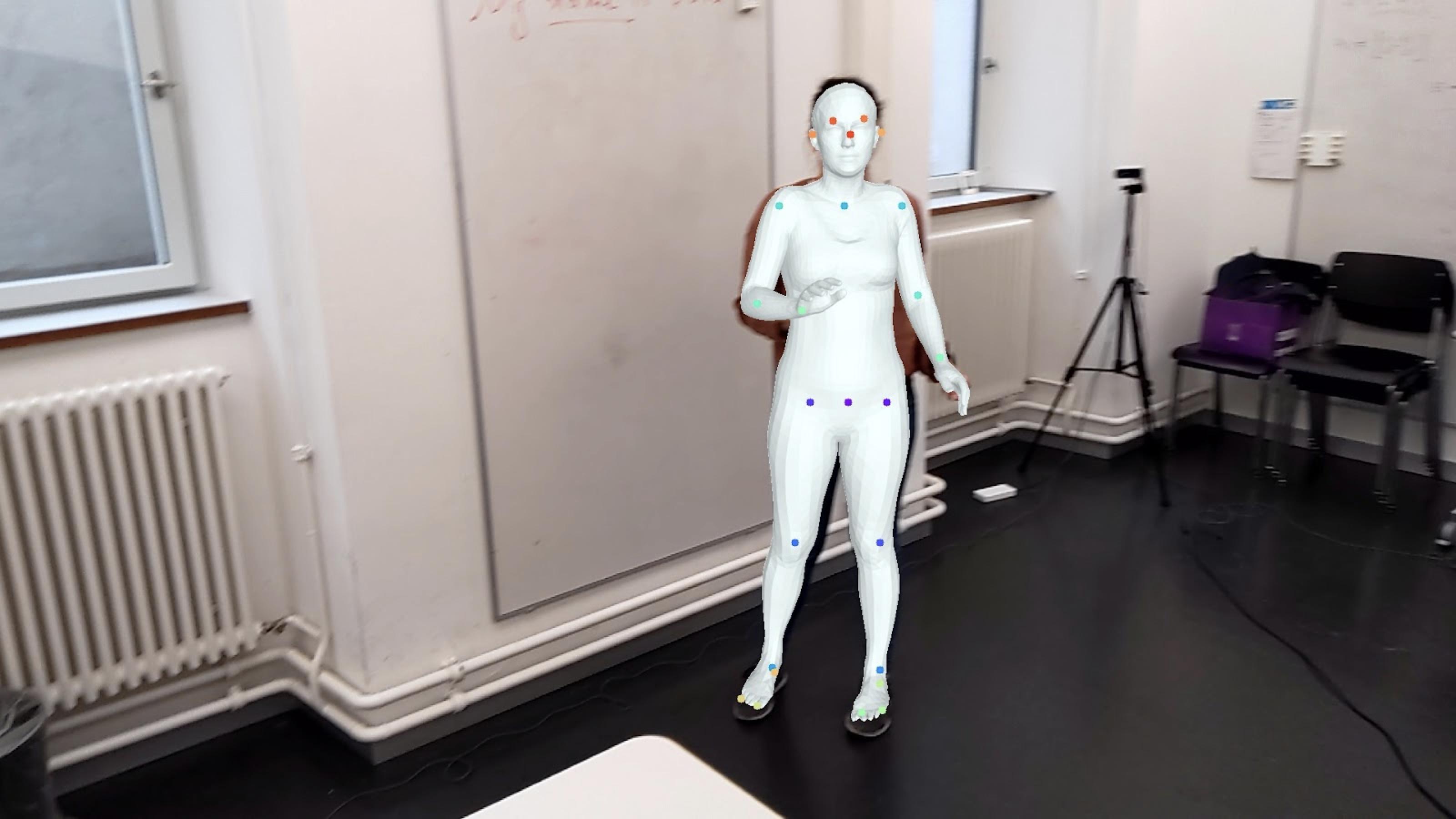}\\
\includegraphics[width=35mm,scale=0.5]{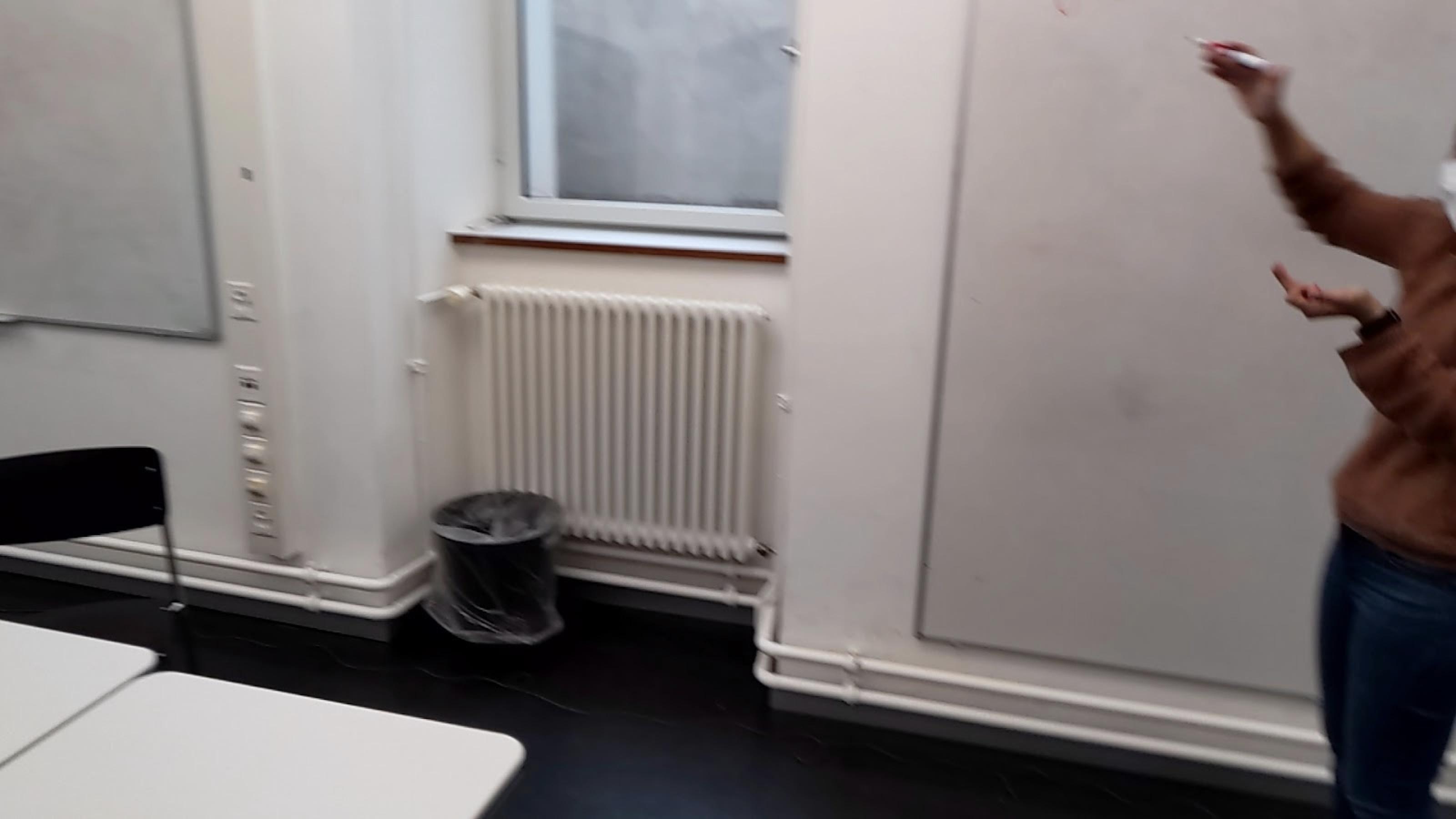} & \includegraphics[width=35mm,scale=0.5]{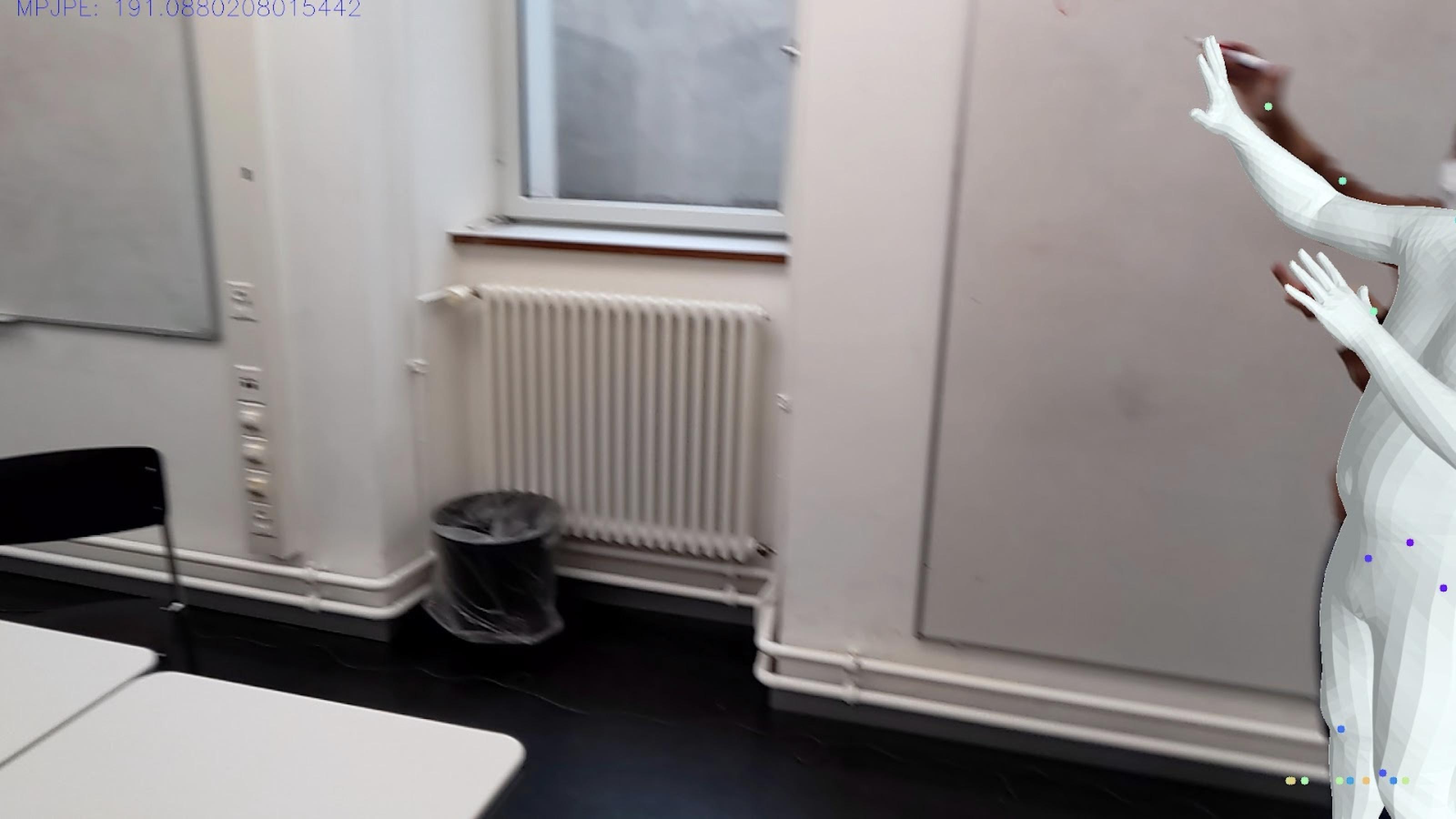} & \includegraphics[width=35mm,scale=0.5]{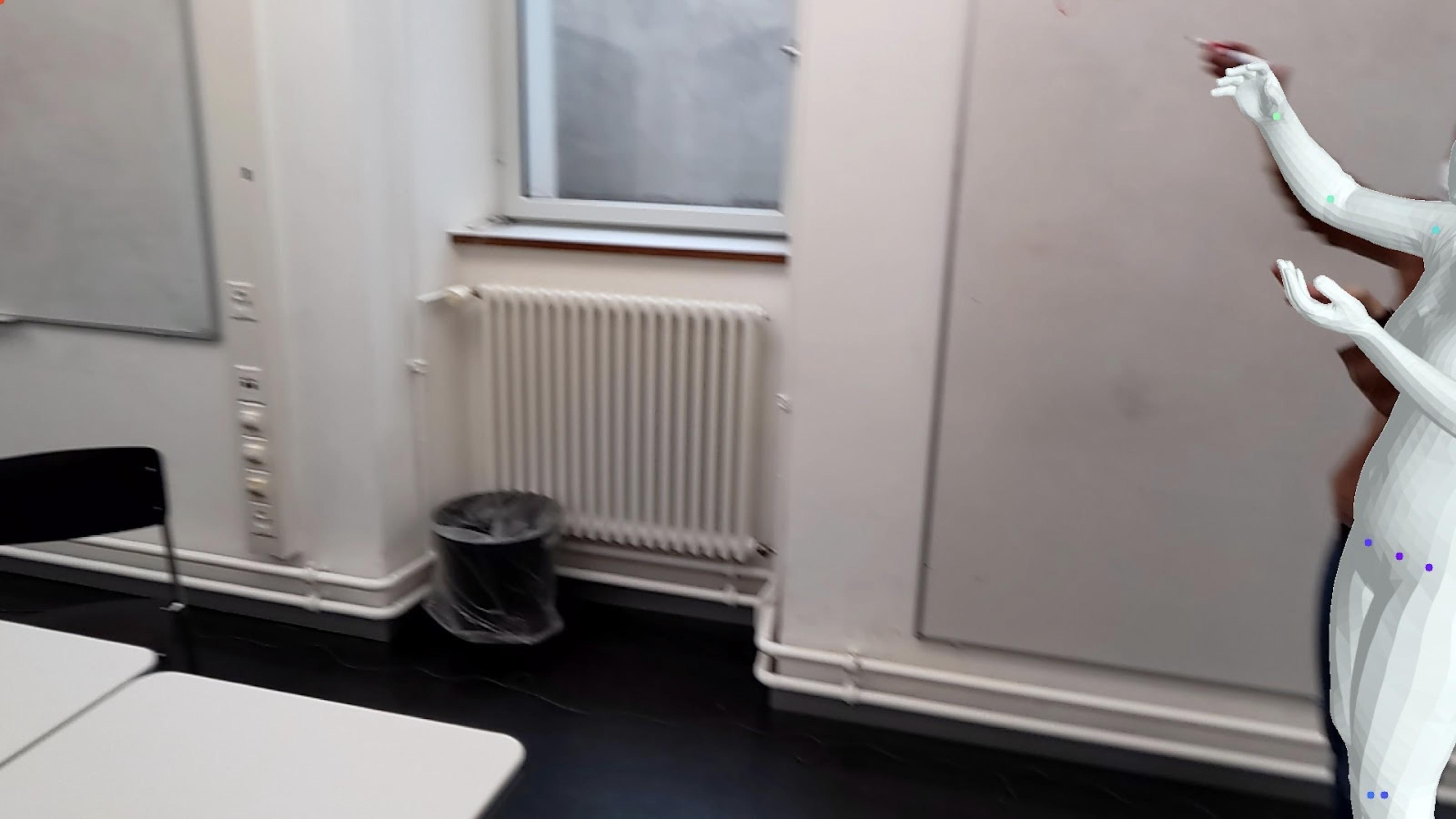}\\
\includegraphics[width=35mm,scale=0.5]{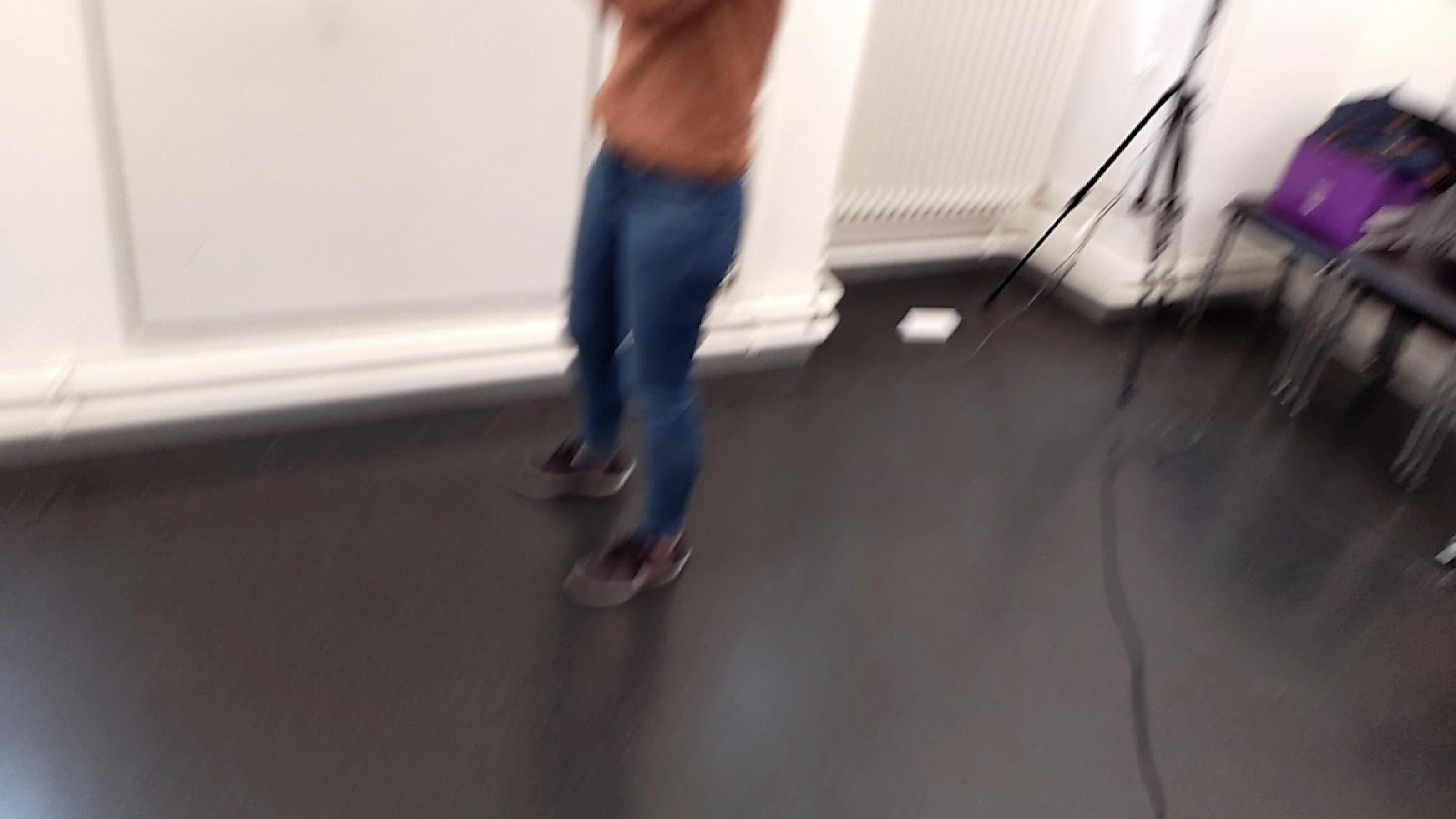} & \includegraphics[width=35mm,scale=0.5]{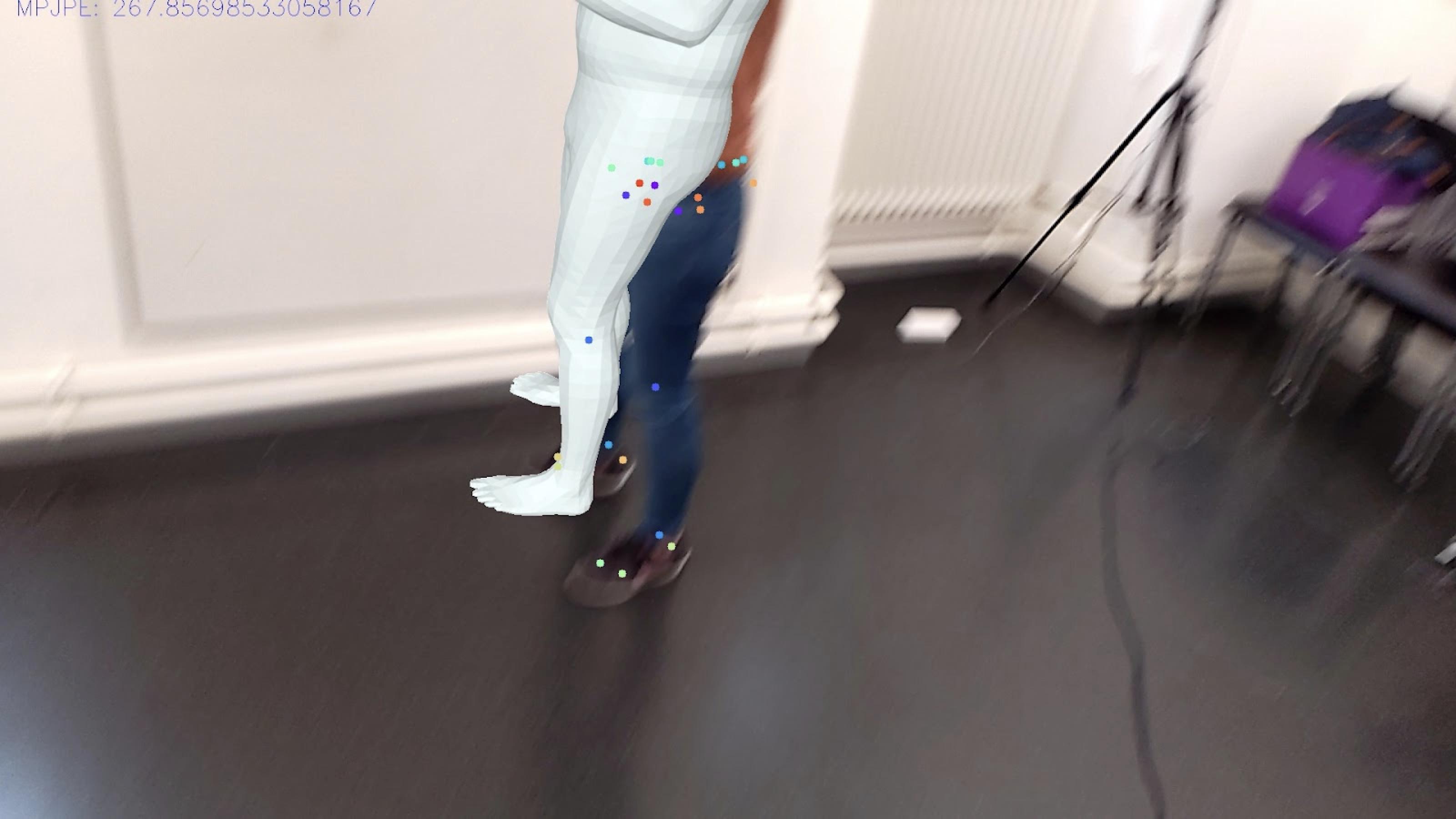} & \includegraphics[width=35mm,scale=0.5]{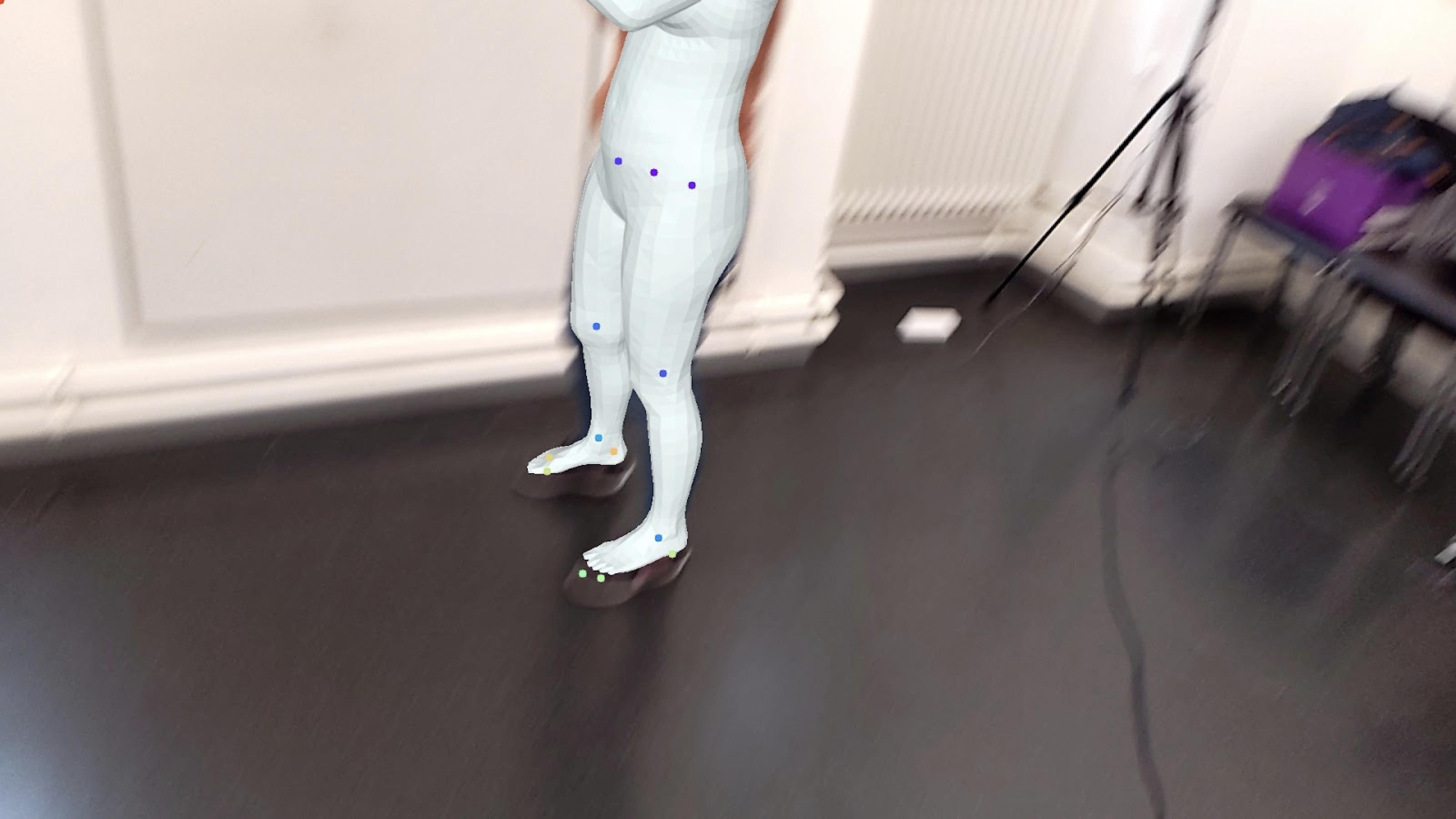}\\
\includegraphics[width=35mm,scale=0.5]{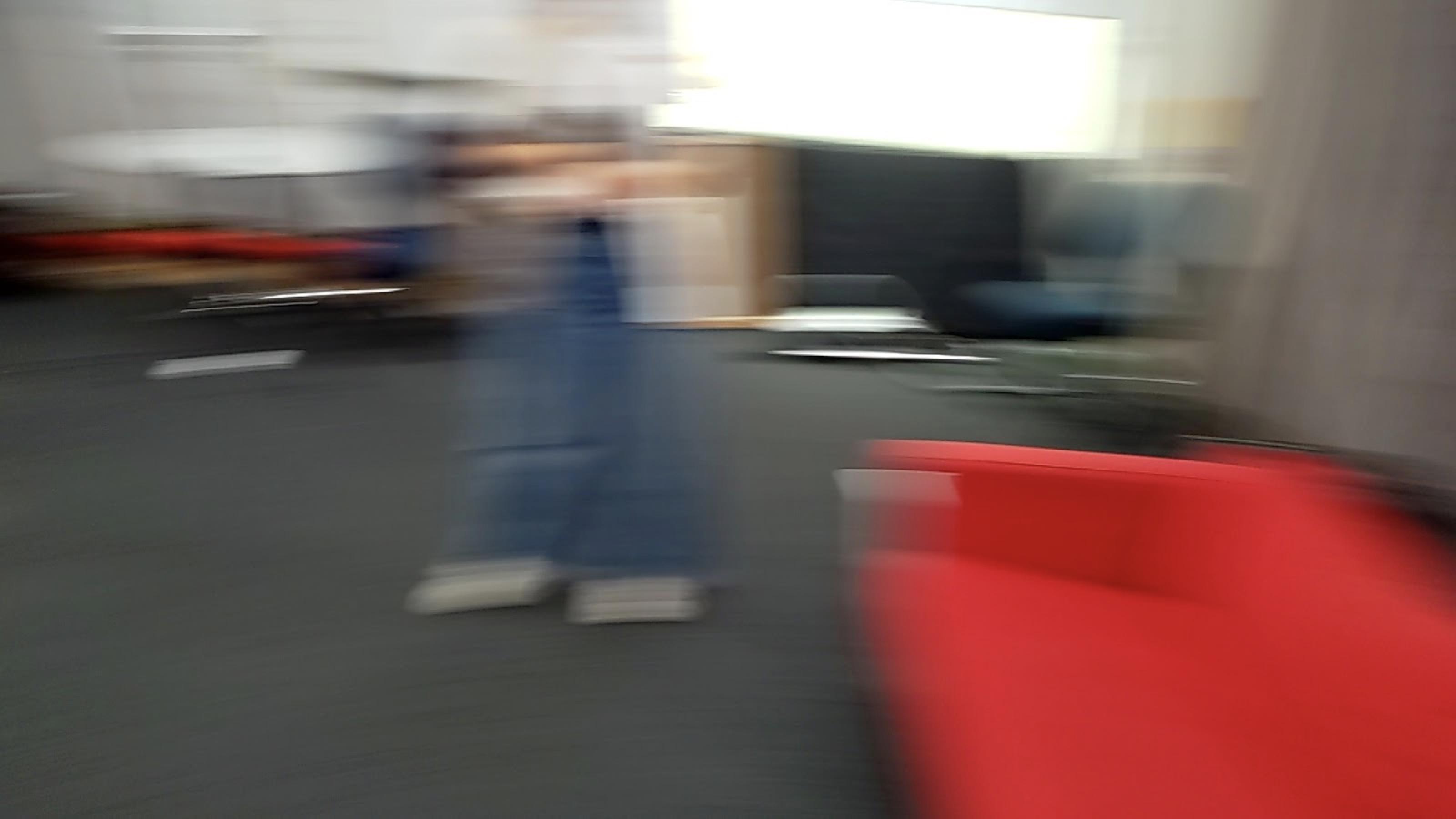} & \includegraphics[width=35mm,scale=0.5]{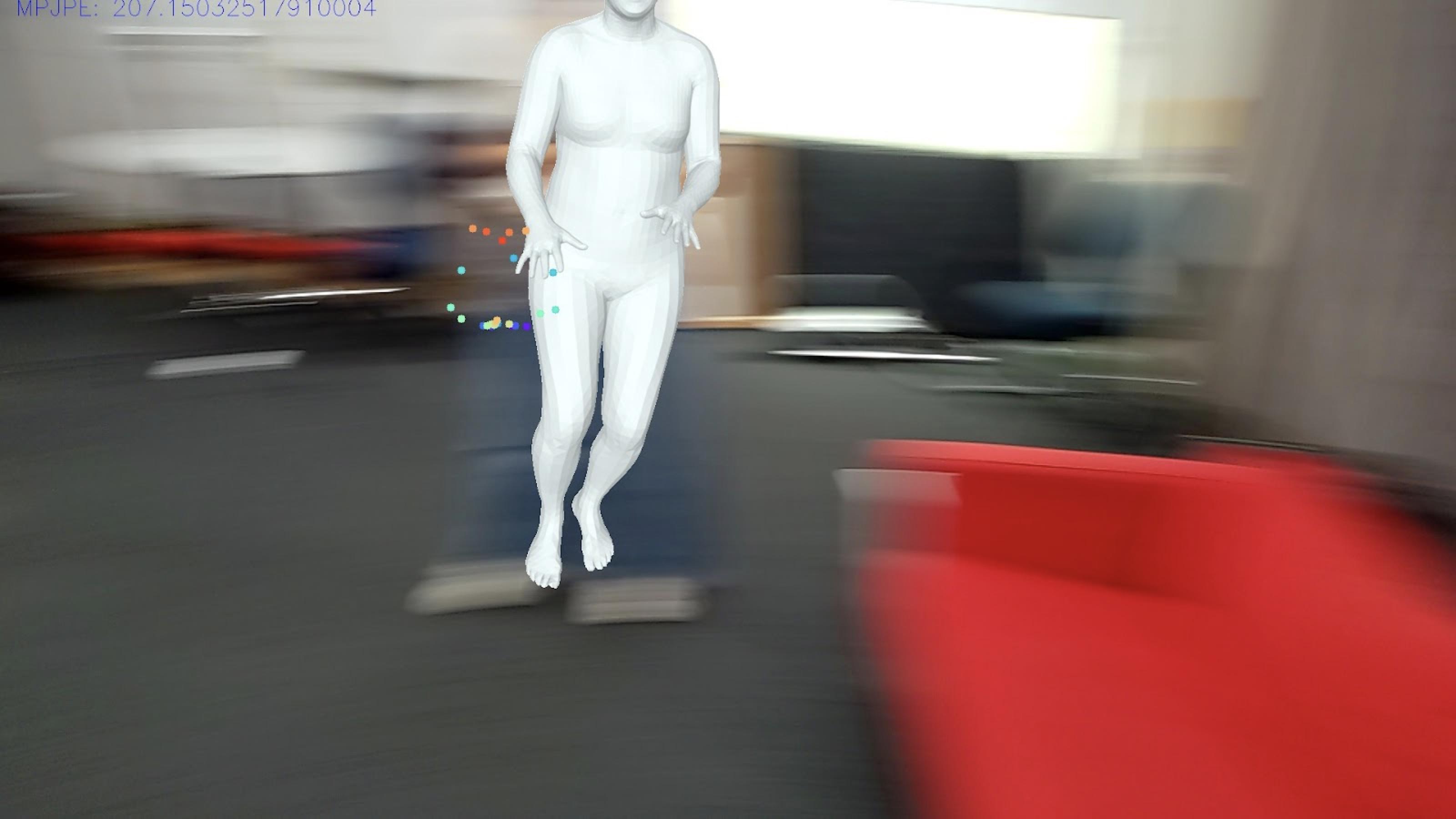} & \includegraphics[width=35mm,scale=0.5]{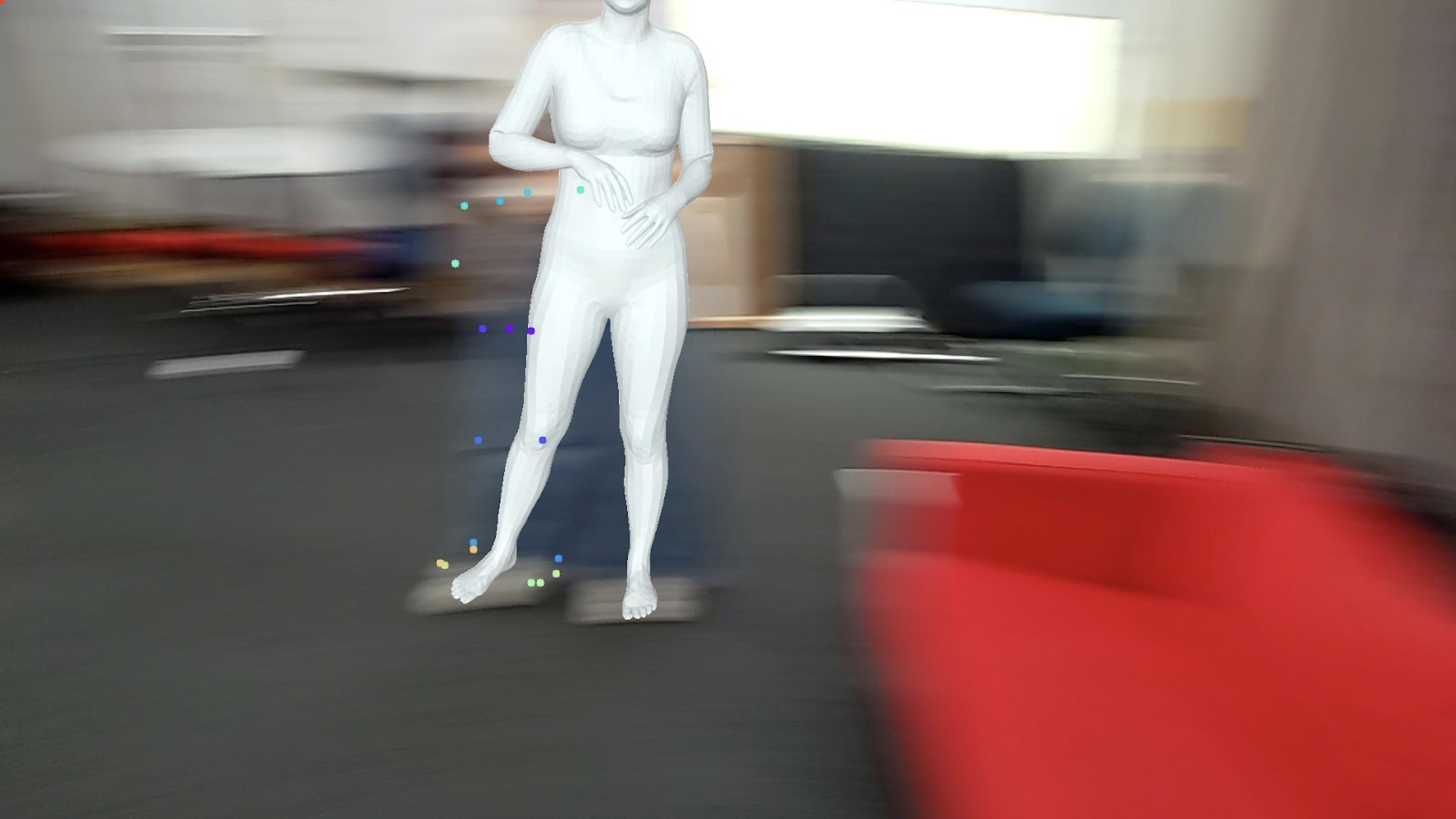}\\
Original Image & Our & GT \\
\end{tabular}
\caption{Visual results of our model compared to GT.}
\label{table:visuals}
\end{figure}

\section{Conclusion}

Due to using multi-scale features and effective pretraining, MEEV achieve MPJPE of 82.30 mm and MPVPE of 92.93 mm for the EgoBody dataset.
For further improvement, exploiting longer temporal information with larger backbone can be favorable.
Besides, exploring to use more various image features can be applicable way to improve the performance.

\clearpage
%
%
\bibliographystyle{splncs04}
\bibliography{main}

\begin{thebibliography}{10}
\providecommand{\url}[1]{\texttt{#1}}
\providecommand{\urlprefix}{URL }
\providecommand{\doi}[1]{https://doi.org/#1}

\bibitem{andriluka14cvpr}
Andriluka, M., Pishchulin, L., Gehler, P., Schiele, B.: 2d human pose
  estimation: New benchmark and state of the art analysis. In: IEEE Conference
  on Computer Vision and Pattern Recognition (CVPR) (June 2014)

\bibitem{resnet}
He, K., Zhang, X., Ren, S., Sun, J.: Deep residual learning for image
  recognition (2015). \doi{10.48550/ARXIV.1512.03385},
  \url{https://arxiv.org/abs/1512.03385}

\bibitem{h36m_pami}
Ionescu, C., Papava, D., Olaru, V., Sminchisescu, C.: Human3.6m: Large scale
  datasets and predictive methods for 3d human sensing in natural environments.
  IEEE Transactions on Pattern Analysis and Machine Intelligence
  \textbf{36}(7),  1325--1339 (jul 2014)

\bibitem{mscoco}
Lin, T.Y., Maire, M., Belongie, S., Bourdev, L., Girshick, R., Hays, J.,
  Perona, P., Ramanan, D., Zitnick, C.L., Dollár, P.: Microsoft coco: Common
  objects in context (2014). \doi{10.48550/ARXIV.1405.0312},
  \url{https://arxiv.org/abs/1405.0312}

\bibitem{liu2022convnet}
Liu, Z., Mao, H., Wu, C.Y., Feichtenhofer, C., Darrell, T., Xie, S.: A convnet
  for the 2020s. Proceedings of the IEEE/CVF Conference on Computer Vision and
  Pattern Recognition (CVPR)  (2022)

\bibitem{SMPL:2015}
Loper, M., Mahmood, N., Romero, J., Pons-Moll, G., Black, M.J.: {SMPL}: A
  skinned multi-person linear model. ACM Trans. Graphics (Proc. SIGGRAPH Asia)
  \textbf{34}(6),  248:1--248:16 (Oct 2015)

\bibitem{vonMarcard2018}
von Marcard, T., Henschel, R., Black, M., Rosenhahn, B., Pons-Moll, G.:
  Recovering accurate 3d human pose in the wild using imus and a moving camera.
  In: European Conference on Computer Vision (ECCV) (sep 2018)

\bibitem{mono-3dhp2017}
Mehta, D., Rhodin, H., Casas, D., Fua, P., Sotnychenko, O., Xu, W., Theobalt,
  C.: Monocular 3d human pose estimation in the wild using improved cnn
  supervision. In: 3D Vision (3DV), 2017 Fifth International Conference on.
  IEEE (2017). \doi{10.1109/3dv.2017.00064}

\bibitem{Moon_2022_CVPRW_Hand4Whole}
Moon, G., Choi, H., Lee, K.M.: Accurate 3d hand pose estimation for whole-body
  3d human mesh estimation. In: Computer Vision and Pattern Recognition
  Workshop (CVPRW) (2022)

\bibitem{Patel:CVPR:2021}
Patel, P., Huang, C.H.P., Tesch, J., Hoffmann, D.T., Tripathi, S., Black, M.J.:
  {AGORA}: Avatars in geography optimized for regression analysis. In:
  Proceedings IEEE/CVF Conf.~on Computer Vision and Pattern Recognition
  ({CVPR}) (Jun 2021)

\bibitem{hrnet}
Wang, J., Sun, K., Cheng, T., Jiang, B., Deng, C., Zhao, Y., Liu, D., Mu, Y.,
  Tan, M., Wang, X., Liu, W., Xiao, B.: Deep high-resolution representation
  learning for visual recognition (2019). \doi{10.48550/ARXIV.1908.07919},
  \url{https://arxiv.org/abs/1908.07919}

\bibitem{Zhang:ECCV:2022}
Zhang, S., Ma, Q., Zhang, Y., Qian, Z., Kwon, T., Pollefeys, M., Bogo, F.,
  Tang, S.: Egobody: Human body shape and motion of interacting people from
  head-mounted devices. In: European conference on computer vision (ECCV) (Oct
  2022)

\end{thebibliography}
\end{document}